\documentclass[letterpaper, 10 pt, conference]{ieeeconf}  %

\IEEEoverridecommandlockouts                              %

\let\labelindent\relax
\usepackage{authblk}
\usepackage{multirow}
\usepackage[usenames,dvipsnames,svgnames,table]{xcolor}
\usepackage{amssymb,latexsym,amsfonts,amsmath}
\usepackage{bm}
\usepackage{dsfont}
\usepackage{mathtools,mathrsfs}
\usepackage{accents}
\usepackage{comment}
\usepackage{keyval}
\usepackage[normalem]{ulem}
\usepackage{graphicx}
\usepackage{float} 

\usepackage{enumitem}

\usepackage{balance}
\usepackage{subcaption} %
\usepackage{caption} %
\usepackage{hyperref}
\usepackage{thmtools}
\usepackage{thm-restate}
\usepackage{multirow}
\usepackage{algorithm}
\usepackage[noend]{algpseudocode}
\usepackage{cite} %

\usepackage[utf8]{inputenc}
\usepackage{tikz}
\usetikzlibrary{positioning}
\usetikzlibrary{arrows.meta}
\usetikzlibrary{shapes,arrows,arrows,positioning,fit}

\usepackage[nolist]{acronym} %

\newacro{ds}[DS]{Dynamical System}
\newacro{lfd}[LfD]{Learning from Demonstration}
\newacro{sos}[SOS]{Sum-Of-Squares}
\newacro{snnds}[S$^2$-NNDS]{Safe and Stable Neural Network Dynamical Systems}
\newacro{pac}[PAC]{Provably Approximately Correct}
\newacro{smt}[SMT]{Satisfiability Modulo Theories}
\newacro{abcds}[ABC-DS]{obstacle Avoidance with Barrier-Certified polynomial
Dynamical Systems}
\newacro{mse}[MSE]{Mean Square Error}
\newacro{}[]{}
\newacro{}[]{}
\newacro{}[]{}
\newacro{}[]{}
\newacro{}[]{}
\newacro{}[]{}
\newacro{}[]{}
\newacro{}[]{}

\definecolor{myLightGreen}{rgb}{0.6, 1.0, 0.6}
\definecolor{myDarkGreen}{rgb}{0.0, 0.5, 0.0}
\definecolor{myPaleGreen}{rgb}{0.8, 0.98, 0.8}
\definecolor{limegreen}{rgb}{0.196, 0.804, 0.196} %
\definecolor{mygreen}{rgb}{0.0, 0.5, 0.0}   %
\definecolor{limegreen}{rgb}{0.196, 0.804, 0.196} %
\definecolor{forestgreen}{rgb}{0.133, 0.545, 0.133} %
\definecolor{seagreen}{rgb}{0.180, 0.545, 0.341} %
\definecolor{springgreen}{rgb}{0.0, 1.0, 0.498} %

\newcommand{\R}{{\mathbb{R}}}

\newcommand{\B}{\mathrm{B}}

\newcommand{\N}{{\mathbb{N}}}

\newcommand{\D}{\mathcal{D}}
\newcommand{\C}{\mathcal{C}}
\newcommand{\Se}{\mathcal{S}}

\newcommand{\bfm}{\mathbf}

\newtheorem{theorem}{Theorem}

\newtheorem{proposition}[theorem]{Proposition}

\newcommand{\eg} {{e.g.,}~} %
\newcommand{\ie} {{i.e.,}~} %

\title{\LARGE \bf
Safe and Stable Neural Network Dynamical Systems\\
for Robot Motion Planning
}

\author{Allen Emmanuel Binny$^{1*}$, Mahathi Anand$^{2*}$, Hugo T.M. Kussaba$^3$, Lingyun Chen$^2$, Shreenabh Agrawal$^4$, \\  Fares J. Abu-Dakka$^5$, Abdalla Swikir$^6$%
\thanks{*The authors contributed equally. $^{1}$A.E. Binny is with the Indian Institute of Technology Kharagpur, $^2$ M. Anand and L. Chen are with the Munich Institute of Robotics and Machine Intelligence, Technical University of Munich, $^3$ H.T.M. Kussaba is with the University of Brasilia, 
$^4$S. Agrawal is with the Carnegie Mellon University, $^5$ F. J. Abu-Dakka is with the Mechanical Engineering Program at New York University Abu Dhabi, 
$^6$A. Swikir is with the Mohamed bin Zayed University of Artificial Intelligence (MBZUAI). This work was partly funded by the German Research Foundation (DFG, Deutsche Forschungsgemeinschaft) as part of Germany’s Excellence Strategy – EXC 2050/1 – Project ID 390696704 – Cluster of Excellence “Centre for Tactile Internet with Human-in-the-Loop” (CeTI) of Technische Universität Dresden, and the Bavarian State Ministry for Economic Affairs, Regional Development and Energy (StMWi) funding of the Lighthouse Initiative KI.FABRIK Bayern Phase 1: Aufbau Infrastruktur.}}

\begin{document}

\bstctlcite{IEEEexample:BSTcontrol}

\maketitle
\thispagestyle{empty}
\pagestyle{empty}

\begin{abstract}
Learning safe and stable robot motions from demonstrations remains a challenge, especially in complex, nonlinear tasks involving dynamic, obstacle-rich environments.
In this paper, we propose Safe and Stable Neural Network Dynamical Systems S$^2$-NNDS, a learning-from-demonstration framework that simultaneously learns expressive neural dynamical systems alongside neural Lyapunov stability and barrier safety certificates. Unlike traditional approaches with restrictive polynomial parameterizations, S$^2$-NNDS leverages neural networks to capture complex robot motions, providing probabilistic guarantees through split conformal prediction in learned certificates. Experimental results in various 2D and 3D datasets---including LASA handwriting and demonstrations recorded kinesthetically from the Franka Emika Panda robot---validate the effectiveness of S$^2$-NNDS in learning robust, safe, and stable motions from potentially unsafe demonstrations. The source code, supplementary material and experiment videos can be accessed via \href{https://github.com/allemmbinn/S2NNDS}{https://github.com/allemmbinn/S2NNDS}.
\end{abstract}

\section{INTRODUCTION}

\ac{lfd} is a paradigm in robotics that teaches robots the essence of a task through multiple demonstrations \cite{BillardCalinonDilmannSchall2008}. This enables a robot to generalize a learned task to new environments, allowing safe navigation under perturbations. A key advantage of \ac{lfd} is its accessibility to end-users without technical expertise, as it requires minimal time and effort to adapt robot's capabilities to diverse environments. 
A prominent approach in \ac{lfd} is to encode demonstrations as stable \ac{ds}~\cite{BillardMirrazaviFigueroa2022book}. This involves recording the robot's end effector or joint positions and velocities and then solving an optimization problem to find a stable \ac{ds} that accurately replicates these demonstrations. Although the demonstrations capture the essence of the task, the inherent stability of \ac{ds} ensures generalization and robust performance under perturbations, enhancing data efficiency by reducing the need for additional demonstrations.

However, a critical limitation of many \ac{lfd} methods is that the optimization problem generally does not consider the obstacles that may be placed in the environment. Consequently, the resulting robot motion may not be safe. This highlights the need to integrate safety constraints directly into the \ac{lfd} framework alongside stability guarantees.

In this paper, we propose a novel framework for learning a \ac{ds} that is not only stable, ensuring convergence to a target equilibrium point under perturbations, but also safe, guaranteeing avoidance of \emph{static obstacles} in the environment. Our learning-based algorithm simultaneously identifies neural network-based \ac{ds} that mimics the provided demonstrations while co-synthesizing formally verified neural stability and safety certificates, characterized by Lyapunov and barrier functions, respectively. The entire \ac{ds} is learned offline, avoiding the need for any real-time modification of \ac{ds}.  

\begin{figure*}[t]
\vspace*{1.0em}%
\centering
\includegraphics[width=\linewidth]{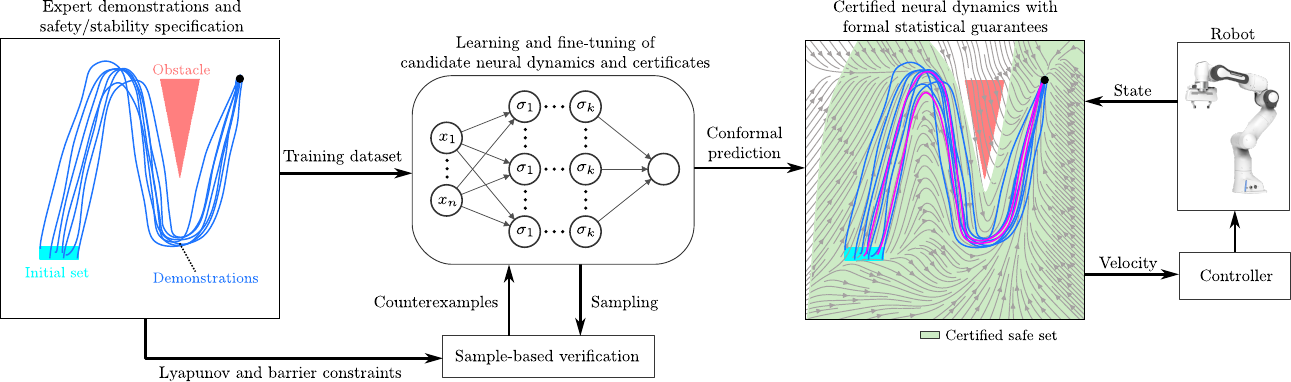}
\caption{Overview of \ac{snnds} Framework.
Neural dynamics are first learned from expert demonstrations and iteratively refined to satisfy Lyapunov and barrier constraints using counterexamples. Verification with conformal prediction then provides formal statistical guarantees on safety and stability.
}
\label{fig:overview}
\vspace{-\baselineskip}
\end{figure*}

\subsection{Related Work}

Dynamical properties like stability and safety can be certified without explicit trajectory computation utilizing the so-called \emph{certificates}. Particularly, \emph{Lyapunov functions} can prove the asymptotic stability of equilibrium points of a system~\cite{khalil_nonlinear_2013}, while \emph{barrier functions} can ensure that a system avoids undesirable states~\cite{Prajna2004safety}.  A common strategy is to synthesize these certificates concurrently with the \ac{ds}~\cite{ds-gmm, figueroa-ds-bayesian, figueroa-ds-gmm, schonger_learning_2024,nawaz_learning_2024}. 

A standard approach for this synthesis is \ac{sos} optimization~\cite{BlekhermanParriloThomas2012-SDPConvexAG}, which reformulates the stability and/or safety constraints as \ac{sos} constraints~\cite{GieslHafstein15-LyapunovComputation, papachristodoulou_lyapunov_sos,clark_verification_2021}. However, \ac{sos} techniques are generally limited to systems with polynomial dynamics and can be overly conservative. In fact, there are globally asymptotically stable polynomial \acp{ds} that lack a polynomial Lyapunov function~\cite{AhmadiKrsticParrilo2011-counterexample}.
Furthermore, high-degree polynomials in \ac{sos} programs are prone to numerical problems, such as floating point errors and poorly conditioned problem formulations, which can undermine the reliability of the results~\cite{roux2018validating}.

\emph{Neural network (NN)-based certificate synthesis} offers a scalable alternative to the limitations of \ac{sos}-based approaches~\cite{abate_neural_2021, AbateAhmedEdwardsGiacobbePeruffo2021-FOSSIL, anand_formally_2023, liu2025physics}. By parameterizing certificates as NNs, these methods take advantage of the universal approximation properties of NNs~\cite{barron_approximation_1994} to find less conservative certificates for more complex motions~\cite{mathiesen2022safety}. However, learning-based certificate synthesis introduces intrinsic errors that can compromise formal guarantees. Thus, formal verification of the learned certificates is necessary, \eg using \ac{smt} solvers~\cite{abate_neural_2021, AbateAhmedEdwardsGiacobbePeruffo2021-FOSSIL, zhao2023formal} or using Lipschitz continuity arguments~\cite{anand_formally_2023}. However, these approaches are either not scalable to deep neural networks or are highly conservative and require dense grid-based sampling of the domains of interest. Recently, \emph{conformal prediction}~\cite{vovk_conditional_2012} has emerged as a promising, sample-efficient technique for providing \ac{pac} guarantees for the formal verification of learning-based algorithms, including applications in safe planning~\cite{lindemann_safe_2023}, formal abstractions~\cite{conf-1}, and reachability~\cite{conf-2}. Conformal prediction was also used in the context of control barrier functions in~\cite{conf-ncbf}. This approach can quantify the validity of stability and safety constraints with high sample efficiency, which positions it as a promising method for the formal verification of learned certificates.

Despite the progress in NN-based certificate learning, few works have explored integrating it with the \ac{lfd} framework. Most existing works in \ac{lfd} either address only stability~\cite{ds-gmm, figueroa-ds-bayesian, figueroa-ds-gmm, Agrawal_learning_2025, boetius2025stable} or consider obstacles in an online setting~\cite{dynamical_obst_avoid,huber_avoidance_2019,huber_avoiding_2022, nawaz_learning_2024}. These approaches modulate \ac{ds} in real-time to deal with dynamic obstacles; however, altering the \ac{ds} can potentially deviate the robot's motions from initial demonstrations. Furthermore, while~\cite{nawaz_learning_2024} considers an NN-based approach to learn safe and stable \ac{ds}, they do not synthesize any certificates and consider them fixed and known, resulting in conservatism. Moreover, their approach is affected by potential conflicts between stability and safety constraints, resulting in compromise between safety and stability. 
Finally, because robots operate under strict real-time control frequency constraints, incorporating online optimization within the control loop significantly increases the computational burden.
 The  most closely related work to our paper is~\cite{schonger_learning_2024}, where the authors proposed \ac{abcds}, an \ac{sos}-based approach to identify safe and stable \ac{ds} against static obstacles.

\subsection{Contributions}
This paper makes the following contributions:
\begin{enumerate}
    \item We propose \ac{snnds}, a novel \emph{offline} approach to generate time-invariant, safe, and stable \acp{ds} directly from demonstrations without requiring online modifications, alleviating the aforementioned challenges. \ac{snnds} jointly learns the \ac{ds} and its corresponding neural certificates, even from unsafe demonstrations. 
    \item Our approach uses expressive neural \emph{network representations} and incorporates \emph{split conformal prediction} to provide probabilistic guarantees on the correctness of the learned stability and safety certificates. An overview is illustrated in Fig.~\ref{fig:overview}.
    \item We validate \ac{snnds} through experiments in 2D and 3D environments, including a publicly available handwriting dataset and kinesthetic demonstrations collected using a Franka Emika Panda robot and compare them with~\cite{schonger_learning_2024}. The results show that our method reliably learns safe and stable motions with minimal restrictions on the obstacle configurations, and is competitive with existing approaches. 
\end{enumerate}
\section{Preliminaries}\label{sec:prelim}
\subsection{Notations}
Let $\R$ and $\N$ denote the sets of real and non-negative integers, respectively. The subscripts specify subsets of $\R$ and $\N$; \eg $\R_{\geq 0}$ is the set of non-negative reals. We use $\R^n$ for a real space of $n$-dimension and $\R^{m \times n}$ for the space of $m \times n$ real matrices.  The cardinality of a finite set $X$ is denoted by $|X|$. 
For a vector (denoted in lowercase) $x \in \R^n$, $x_i$ is its $i^{\text{th}}$ element, $1 \leq i \leq n$. The Euclidean norm of $x$ is $\|x\|$.  For a differentiable function $f: X \rightarrow Y,$ where sets $X, Y \subseteq \R^n$, and $\nabla f$ denotes its gradient, \ie $\nabla f(\mathbf{x})=\left(\partial f / \partial x_{1}(x), \ldots, \partial f / \partial x_{n}(x)\right)^{\top}$. The notation $ \circ $ and $\mathrm{id}$ denote the composition of functions and the identity function, respectively. The indicator function $\mathbb{I}_X(x)$ is $1$ if $x \in X$ and $0$ otherwise. The ceiling and floor functions are $\lceil \cdot \rceil$ and $\lfloor \cdot \rfloor$, respectively. The regularized incomplete Beta function is
$\mathcal{I}_c(a,b) = \frac{\int_{0}^c t^{a-1}(1-t)^{b-1} \mathrm{d}t}{B(a,b)}$, where ${B(a,b)=\int_{0}^1 t^{a-1}(1-t)^{b-1} \mathrm{d}t}$ is the beta function. 
Lastly, $\bfm x$ (denoted in bold letters) represents both continuous trajectories and distinct sequences. The distinction will become clear from the context. 

\subsection{\texorpdfstring{\protect\ac{ds} for Motion Planning}{DS for Motion Planning}}

The \ac{lfd} framework for robot motion planning aims to learn a \ac{ds} described by the following differential equation
\vspace{-0.25em}
\begin{equation}
\dot{\bfm x}(t) = f(\bfm x(t)),  \label{eq:ds}
\end{equation}
\vspace{-0.25em}
where $\bfm x(t) \in X$ is the state variable (\eg Cartesian position of the robot's end-effector within its workspace $X$) at $t \in \R_{\geq 0}$. $f: X \rightarrow X$ is a continuous non-linear function that describes the state velocity of the robot. We assume that \eqref{eq:ds} has a unique solution for every initial condition $\bfm{x}(0) \in X$. 

Given a set of $N$ demonstrations, $\mathcal{D} = \{ (x_{ij}, \dot{x}_{ij}) \}_{i=1, j=1}^{N, M}$, where each demonstration consists of $M$ time-discretized positions $\bfm x_i = \big (\bfm x_i(t_1), \bfm x_i(t_2), \ldots, \bfm x_i(t_M) \big)$ and corresponding velocities $\dot{\bfm x}_i = \big (\dot{\bfm x}_i(t_1), \dot{\bfm x}_i(t_2),  \ldots, \dot{\bfm x}_i(t_M) \big)$,  $t_0 = 0$, $t_M > 0$, $i \in \{1,\ldots,N\}$.
The goal of this work is to learn a function $f$ such that the resultant \ac{ds} in~\eqref{eq:ds} not only approximates the demonstrations $\D$ but also converges to a goal state $x^\star$ while avoiding obstacles in $X$. Under nominal conditions, \ie in the absence of external disturbances or obstacles, one can learn a \ac{ds} simply by minimizing the \ac{mse} between $f$ and $\dot{\bfm x}$, \ie

\vspace{-0.3em}
\begin{equation}
\min_f \frac{1}{MN} \sum_{i=1}^N \sum_{i=1}^M \|\dot x_{ij} - f(x_{ij}) \|^2
\label{eq:mse}
\end{equation}
\vspace{-2pt}
However, since $f$ is learned using only a finite number of demonstrations, the \ac{ds} is not robust, \ie under any external perturbations during deployment, the trajectory may not converge to the required goal $x^\star$. In addition, since obstacles are not considered during training, the resulting trajectories may not be safe. Therefore, to ensure robust convergence and obstacle avoidance, we must incorporate formal certificates of stability and safety. It is well-established that the stability of a \ac{ds} can be certified using Lyapunov functions~\cite{khalil_nonlinear_2013}, which are formulated below.

\begin{proposition}[Asymptotic Stability] ~\label{def:stab}
Consider the \ac{ds} in~\eqref{eq:ds} with an isolated equilibrium point $x^\star \in X$. A continuously differentiable real-valued function $V: X \rightarrow \R$ is a \emph{Lyapunov function}, where $V(x^\star) = 0$, $\dot{V}(x^\star) = 0$, if for all $x \in X \setminus \{x^\star\}$ the following conditions are satisfied:
\vspace{-2pt}
\begin{equation}
V(x) > 0, \quad \dot V(x) = \nabla V(x)^T f(x) < 0. ~\label{eq:lyap}
\end{equation}
\vspace{-2pt}
The existence of a Lyapunov function guarantees that the system is locally asymptotically stable within $X$ and $\exists X_i \subseteq X \text{ s.t. } \forall \bfm{x}(0) \in X_i, \exists t \in \R_{\geq 0}$ with $\bfm{x}(t) = x^\star$.
\end{proposition}
The proof is standard and a direct application of the Lyapunov theory~\cite[Theorem 4.1]{khalil_nonlinear_2013}, and is thus omitted.
Similarly, barrier functions were first introduced in~\cite{Prajna2004safety} to certify the safety of \acp{ds}. 

\begin{proposition}[Safety] ~\label{def:safe}
Consider the \ac{ds} in~\eqref{eq:ds} with an initial set $X_0 \subseteq X$, \ie $\bfm{x}(0) \in X_0$ and an unsafe set ${X_u \subseteq X}$.  A continuously differentiable real-valued function $\B: X \rightarrow \R$ is a \emph{barrier function} if $\exists\ \varepsilon > 0$ such that:
\begin{alignat}{2}
&  \B(x) \leq 0, \quad  \quad && \forall x \in X_0, ~\label{eq:bar1} \\ 
& \B(x) > 0, \quad \quad && \forall x \in X_u,~\label{eq:bar2} \\
& \nabla B(x)^\top f(x) \leq 0, \  && \forall x \in \{x \in X\mid |B(x)| \leq \varepsilon \}. ~\label{eq:bar3}
\end{alignat}
The existence of a barrier function guarantees that the system is safe, i.e., $\bfm{x}(t) \notin X_u$ for all $t \in [0,T]$ for any $T \in \N$.
\end{proposition}
The proof of Proposition~\ref{def:safe} can be found in~\cite{schonger_learning_2024}.

Learning a \ac{ds} while guaranteeing stability and safety can, therefore, be expressed by minimizing the \ac{mse} in~\eqref{eq:mse} subject to the stability and safety conditions defined by Propositions~\eqref{eq:lyap}-\eqref{eq:bar3}. 
Specifically, this can be formulated as follows\footnote{Without loss of generality, we assume $x^\star = 0$, which can be achieved by translating the coordinate system. We also assume the demonstrations end at the equilibrium, \ie $(\bm{x}_{iM}, \dot{\bm{x}}_{iM}) = (0, 0)$.}:
\vspace{-5pt}
\begin{alignat}{2} 
& \min_f \frac{1}{MN} &&  \sum_{i=1}^{N} \sum_{j=1}^M \| \dot x_{ij} - f(x_{ij}) \|^2  \nonumber \\
& s.t. && f(0) = 0, \nonumber \\
& \quad && \text{conditions from Prop.~\eqref{eq:lyap}--\eqref{eq:bar3} hold}. \label{eq:mse_ss}
\vspace{-\baselineskip}
\end{alignat}
\vspace{-15pt}

Finding a solution to~\eqref{eq:mse_ss} is computationally intractable, since it requires optimizing infinite-dimensional function spaces to find functions $f, V,$ and $\B$. In practice, one needs to obtain suitable parameterizations of the functions. Conventional approaches consider linear parameter-varying dynamics with quadratic Lyapunov functions~\cite{ds-gmm, figueroa-ds-bayesian, figueroa-ds-gmm} or polynomial dynamics with polynomial certificate functions~\cite{schonger_learning_2024}. However, these methods have significant drawbacks. The rigid structure of these parameterizations limits the flexibility needed to represent highly nonlinear and complex motions. They also lack the versatility to handle obstacles with complex shapes, \eg~\cite{schonger_learning_2024} only considers semi-algebraic sets. Furthermore, these parameterizations often lead to highly complex, possibly nonconvex, optimization problems that are computationally demanding to solve. To alleviate these challenges, we propose a neural network parameterization of the above optimization problem. We characterize the functions $f, V,$ and $\B$ as NNs and learn them simultaneously by constructing an appropriate loss functions to enforce~\eqref{eq:mse_ss}. 

\section{Proposed Methodology}
\label{sec:method}

In this section, we introduce our \acf{snnds} framework to solve the optimization problem in~\eqref{eq:mse_ss}. The core idea is to represent the \ac{ds} and the certificate functions using NNs and train them concurrently using carefully designed loss functions. Once the networks are trained, we use conformal prediction to validate the learned certificates. 

\subsection{Neural Network Parameterizations}
\label{subsec:nn}

We parameterize the function $f$ in~\eqref{eq:ds} with a fully connected neural network of $K$ layers, denoted $f_\theta(x)$. The NN consists of $n_0 = n$ inputs, $n_k = n$ outputs, and $K-2$ hidden layers, with each hidden layer $i \in \{1, \ldots, K-2\}$ containing $n_i$ hidden neurons. More precisely, consider
\vspace{-2pt}
\begin{equation} \label{eq:nn_param}
f_\theta(x) = l_{K-1} \circ \ldots \circ  l_0 (x),
\end{equation}
\vspace{-2pt}
where each layer $l_i: \R^{n_i} \rightarrow \R^{n_{i+1}}$ is defined as
\vspace{-2pt}
\begin{equation*}
l_i(x) = \sigma_i(W_ix + b_i), \quad \forall i\in \{0, \ldots, K-1\}.
\end{equation*}
\vspace{-2pt}
Here, $W_i \in \R^{n_{i+1} \times n_i}$ and $b_i \in \R^{n_{i+1}}$ are the weight matrix and the bias vector, respectively, collectively denoted by the parameters $\theta = [W; b]$. $\sigma_i: \R^{n_{i+1}} \rightarrow \R^{n_{i+1}}$ represents the element-wise application of an appropriate activation function (\eg $\mathrm{ReLU}$, $\mathrm{Tanh}$, $\mathrm{ELU}$, etc.). To ensure $f_\theta(0) = 0$ as required in~\eqref{eq:mse_ss}, we set all bias terms $b_i = 0$, $\forall i \in \{0, \ldots, K-1\}$. 

The certificate functions $V$ and $B$ are also parameterized as NNs, $V_{\theta'}$ and $\B_{\theta''}$, similarly to~\eqref{eq:nn_param}, with parameters $\theta'$ and $\theta''$, respectively. However, some distinctions are made to enforce certain properties on these functions. For example, to ensure $V_{\theta'}(0) = 0$ we set the bias terms in $V_{\theta'}$ to $0$ according to Proposition~\ref{def:stab}. No such restriction is placed on $\B_{\theta''}$. Moreover, to ensure continuous differentiability of the certificates (as required for conditions~\eqref{eq:lyap} or~\eqref{eq:bar3}), differentiable activation functions are considered for both
$V_{\theta'}$ and $\B_{\theta''}$. Finally, we set $\sigma_{K-1} = \mathrm{id}$ for these networks.

To train these neural networks, we define a set of loss functions. The primary loss is the \ac{mse} loss from~\eqref{eq:mse}, augmented by the incorporation of the constraints~\eqref{eq:lyap}--\eqref{eq:bar3}. Given a demonstration set $\D$ consisting of $|\D| = MN$ elements of robot's positions and velocities $(x, \dot x) \in \D$ (possibly shuffled), the loss function corresponding to the objective function is formally defined as the \ac{mse} between the demonstrated velocities $\dot x$ and the predicted velocities of the neural network function $f_\theta (x)$:
\vspace{-2pt}
\begin{equation} \label{eq:lmse}
 L_{MSE}(\theta) = \frac{1}{|\D|} \sum_{(x,\dot x) \in \D} \|\dot x - f_\theta(x) \|^2.
\end{equation}
\vspace{-2pt}
To construct a candidate Lyapunov function $V_{\theta'}$, we first sample a set of i.i.d samples from the workspace $X$ as $\Se = \{x_i \in X \mid i \in \{1, \ldots, S\}\}$. We then define a hinge-like loss using a leaky $\mathrm{ReLU}$ function in~\cite{abate_neural_2021}. Then, the loss function corresponding to condition~\eqref{eq:lyap} is obtained as
\vspace{-5pt}
\begin{align} 
L_{lyap}(\theta, \theta') & =  \frac{1}{|\Se|} \sum_{x \in \Se} \big(\lambda_{l1} \mathrm{LR}_\alpha(\delta_{l1} - V_{\theta'}(x)) \nonumber \\ & + \lambda_{l2} \mathrm{LR}_\alpha( \nabla V_{\theta'}(x) f_\theta(x) - \delta_{l2}), \label{eq:llyap}
\end{align}
\vspace{-2pt}
where $\mathrm{LR}_\alpha$ is the leaky ReLU function parameterized by $\alpha \geq 0$, $\lambda_{l1},\lambda_{l2}$ are the loss weights and $\delta_{l1},\delta_{l2}$ are small positive tolerances for numerical stability. This loss $L_{lyap}(\theta, \theta')$ penalizes violations of the Lyapunov conditions~\eqref{eq:lyap} while promoting them to take on values where~\eqref{eq:lyap} is satisfied with a large margin. 

Similarly, for the barrier function $\B_{\theta''}$, we sample points from the initial $\Se_0 = \Se \cap X_0$ and the unsafe set ${\Se_u = \Se \cap X_u}$. Moreover, to enforce condition~\eqref{eq:bar3}, an additional term is included in the loss function\footnote{The last loss term is applied over the entire domain $X$ rather than just the region $\{x \in X \mid |B(x)| \leq \varepsilon \}$ to enhance robustness during training.}. Consequently, the loss function for the barrier certificate is defined as
\vspace{-5pt}
\begin{align} \label{eq:lbar}
\hspace{-0.5em} L_{bar}(\theta, \theta'')  & = \frac{1}{|\Se_0|} \sum_{x \in \Se_0} \lambda_{b2} \mathrm{LR}_\alpha (\B_{\theta''}(x) \nonumber  + \delta_{b2})  \\ & + \frac{1}{|\Se_u|}\sum_{x \in \Se_u} \lambda_{b3} \mathrm{LR}_\alpha (\delta_{b3} - \B_{\theta''}(x)) \nonumber \\ & + \frac{1}{|\Se|} \sum_{x \in \Se}\lambda_{b1} \mathrm{LR}_\alpha( \nabla \B_{\theta''}(x) f_\theta(x) - \delta_{b1}),
\end{align}
where $\lambda_{b1}, \lambda_{b2}, \lambda_{b3}$ represent the weighting coefficients for each term, and $\delta_{b1}, \delta_{b2}, \delta_{b3}$ are the corresponding tolerance parameters. Note that these weights, tolerance parameters, and $\alpha$ are hyperparameters that must be predetermined. 

Minimizing the loss functions described in~\eqref{eq:lmse}--\eqref{eq:lbar} aims to yield a suitable \ac{ds} along with candidate stability and safety certificates for the optimization problem~\eqref{eq:mse_ss}. However, minimization of losses such as~\eqref{eq:lyap} and~\eqref{eq:lbar} does not inherently guarantee the satisfaction of conditions~\eqref{eq:lyap}--\eqref{eq:bar3} throughout the entire continuous domain $X$. Furthermore, even over the sampled dataset, strict satisfaction guarantees are not achieved, as positive loss terms can be canceled out by larger negative terms. Consequently, the functions obtained through this training process offer neither empirical nor formal guarantees on their validity. To address this limitation, we propose to leverage conformal prediction techniques to obtain formal \ac{pac}-like guarantees on the validity of safety and stability certificates.

\subsection{Validation of Certificates via Conformal Prediction}

Conformal prediction~\cite{conf_predic_intro} is an uncertainty quantification technique that is predominantly used in the context of machine learning to obtain statistically rigorous confidence intervals for model predictions. Its core principle involves assigning nonconformity scores to a calibration set of independently drawn data samples. These scores are then used to obtain reliable predictions for new test data drawn from the same underlying distribution. In this paper, inspired by~\cite{conf-1,conf-2,conf-ncbf}, we propose a split conformal prediction-based algorithm. This algorithm aims to establish probabilistic lower bounds on the satisfaction of conditions~\eqref{eq:lyap}--\eqref{eq:bar3} by the learned candidate certificates $V_{\theta'}$ and $\B_{\theta''}$ across the entire workspace $X$. 

To formalize this, consider a learned \ac{ds} $f_{\theta}$ as in~\eqref{eq:ds}, alongside the certificate functions $V_{\theta'}$ and $\B_{\theta''}$. We define a calibration set $\mathcal{C} = \{ x_i \in X \mid i \in \{1, \ldots, N_{ver}\}$ consisting of $N_{ver}$ i.i.d samples. For each $x \in \mathcal{C}$, we compute a nonconformity score function as $s(x) = \max_{q \in \{1, \ldots, 5\}} \rho_q(x)$, which is defined as the maximum violation across all relevant conditions; where \(\rho_q(x)\) represents the violation of the \(q\)-th condition. Specifically, $\rho_1(x), \rho_2(x)$ correspond to the stability conditions in~\eqref{eq:lyap} formulated such that $\rho_q(x) \leq 0$ indicates satisfaction. Similarly, $\rho_q(x)$ for $q \in \{3,4,5\}$ can be expressed in an analogous form to represent violations of the barrier conditions~\eqref{eq:bar3}. For instance, $\rho_3(x) = \B(x)\mathbb{I}_{X_0}(x)$ could represent a violation related to the initial set. A higher value of $\rho_q(x)$ directly signifies a greater violation of the corresponding condition~\eqref{eq:lyap}--\eqref{eq:bar3}. Consequently, obtaining $s(x) \leq 0$ for any randomly drawn test point $x \in X$ would rigorously validate the certificate functions. Building upon this intuition, we present the following theorem, adapted from~\cite{conf-2, conf-ncbf}, which provides a quantifiable measure of satisfaction for conditions~\eqref{eq:lyap}--\eqref{eq:bar3}. 

\begin{theorem}[Verification via Conformal Prediction]
\label{thm:CP}
Let $f_{\theta}$ be a learned \ac{ds} and $V_{\theta'}$ and $\B_{\theta''}$ be the corresponding candidate Lyapunov and barrier functions, respectively. Consider a calibration set $\mathcal{C} = \{x_1, \ldots, x_{N_{ver}}\}$ consisting of  $N_{ver}$ i.i.d. samples. Given confidence levels $\alpha, \beta, \epsilon \in (0,1)$ such that the following condition holds for the regularized incomplete beta function $\mathcal{I}_{1-\epsilon}(N_{ver} - l + 1, l) \leq \beta$, where $l = \lfloor (N_{ver})(\alpha) \rfloor$. Then, with a confidence of at least $1-\beta$, the probability that a randomly drawn test point $x \in X$ satisfies $s(x) \leq p$ is at least $1-\epsilon$:
\begin{equation} \label{eq:cpprob}
\mathbb{P}_{x \in X} (s(x) \leq p) \geq 1-\epsilon,
\end{equation}
where $p$ is the quantile, $\frac{\lceil (N_{ver+1} +1)(1-\alpha)\rceil}{N_{ver}}^{th}$, of the scores $s(x), \forall x \in \mathcal{C}$. 
\end{theorem}
The proof of Theorem~\ref{thm:CP} is analogous to the methodology presented in~\cite{conf-2}. $p \leq 0$ indicates the satisfaction of conditions~\eqref{eq:lyap}--\eqref{eq:bar3} for some sufficiently high confidence levels $(1-\beta)$ and $(1-\epsilon)$, while $p > 0$ indicates the presence of safety violations~\cite{conf-ncbf}. Therefore, to obtain a statistically significant guarantee on the validity of the safety and stability certificates, it is imperative to ensure that $p \leq 0$. For a more intuitive explanation of formal verification via conformal prediction, we refer the readers to~\cite{conf-2}. The general algorithm to verify the validity of these safety conditions is detailed in Algorithm~\ref{alg:cp}.

\vspace*{-\topskip}  %
\begin{algorithm}[H]
    \centering
    \caption{Certificate Verification using Conformal Prediction}
    \label{alg:cp}
    \begin{algorithmic}
    \Function{CP}{$f_\theta, V_{\theta'}, \B_{\theta''}, N_{ver}, \epsilon$}
    \State verified $\gets$ \textbf{false}
    \State $\C \gets$ Sample $N_{ver}$ i.i.d states from $X$
    \State $S = \emptyset$
    \For{$x \in \C$}
    \State S.insert(s(x))
    \EndFor
    \State $S \gets$ Sort $S$ in non-decreasing order
    \State $\alpha, \beta \gets$ Solve $\mathcal{I}_{1-\epsilon}(N_{ver} - l + 1, l) \leq \beta$ 
    \State $p \gets \hspace{-0.3em} \frac{\lceil (N_{ver+1} +1)(1-\alpha)\rceil}{N_{ver}}^{th}$ quantile of $S$
    \If{$p \leq 0$}
    \State verified $\gets$ \textbf{true}
    \EndIf
    \State \textbf{return} $p, 1-\beta$, verified
    \EndFunction
    \end{algorithmic}
\end{algorithm}
\vspace{-1.5em}

\subsection{\texorpdfstring{The \ac{snnds} Algorithm}{The S2-NNDS Algorithm}}

The \ac{snnds} algorithm for learning safe and stable \ac{ds} is structured in two distinct phases: training and verification. The training phase begins with an initial training of $f_\theta$ using the \ac{mse} loss~\eqref{eq:lmse} on the demonstration dataset $\D$. However, this initial dynamics model does not inherently guarantee safety or stability, as the corresponding certificates may not yet exist. Consequently, $f_\theta$ is further fine-tuned within a counterexample guided scheme, concurrently learning the Lyapunov and barrier certificate functions.  This is done by training networks with respect to composite loss functions~\eqref{eq:lmse}--\eqref{eq:lbar} with the aim of achieving verification guarantees based on samples with minimal data. Initially, a small finite sample dataset $\Se \subseteq X$ is established, and NNs $f_{\theta}, V_{\theta'}$ and $\B_{\theta''}$ are trained simultaneously for a specified number of epochs. Following this, a few counterexamples from a randomly generated larger finite set $\Se_{cex} \subset X$ that violate conditions~\eqref{eq:lyap}--\eqref{eq:bar3} are iteratively added to $\Se$ and the networks are re-trained until no further counterexamples are found. At this stage, only statistical verification guarantees are available for the $|\Se_{cex}|$ samples. To obtain stronger formal statistical guarantees on the trained candidate certificates, Algorithm~\ref{alg:cp} is applied as a posterior verification step. The general algorithm is formalized in Algorithm~\ref{alg:nnds}.

\begin{figure}[!t]%
\vspace*{-0.71em}%
\begin{algorithm}[H]
    \centering
    \caption{The \ac{snnds} Algorithm}\label{alg:nnds}
    \begin{algorithmic}
        \Require \parbox[t]{\dimexpr\linewidth-\algorithmicindent}{%
            $X, X_0, X_u, \D, \Se,$\\
            $N_{cex}, N_{ver}, \mathrm{epochs}, \mathrm{iters}, \epsilon$
        } \vspace{0.001em} 
        \State $\Se_0, \Se_u \gets \mathcal{S} \cap X_0, \mathcal{S} \cap X_u$
        \State Train $f_\theta$ subject to loss~\eqref{eq:lmse}
        \State Initialize  $V_{\theta'}, \B_{\theta''}$
        \For{$i \leq iters$}
        \For{$j \leq epochs$}
         \State Train $f_{\theta}, V_{\theta'}, \B_{\theta''}$ w.rt. losses~\eqref{eq:lmse}--\eqref{eq:lbar}
         \EndFor 
        \State $\mathcal{S}_{cex} \gets$ Sample $N_{cex}$ i.i.d states from $X$
        \State $no_{cex}, \mathcal{S}_{viol} \gets$ number and set of counterexamples violating conditions~\eqref{eq:lyap}--\eqref{eq:bar3}
        \If{$no_{cex} = 0$}
        \State \textbf{break}
        \Else
        \State $\mathcal{S}_0 \gets \mathcal{S}_0 \wedge \mathcal{S}_{viol}, \mathcal{S}_u \gets \mathcal{S}_u \wedge \mathcal{S}_{viol}, \mathcal{S} \gets \mathcal{S} \wedge \mathcal{S}_{viol}$
        \EndIf
        \EndFor
        \If{$no_{cex} = 0$}
        \State $p, 1-\beta, \text{verified} \gets \mathrm{CP}(f_\theta, V_{\theta'}, \B_{\theta''},$ $ N_{ver}, \epsilon$)
        \State \textbf{return} $f_{\theta}, V_{\theta'}, \B_{\theta''}, p, 1-\beta, \text{verified}$
        \EndIf
        \textbf{return none}
    \end{algorithmic}
\end{algorithm}
\vspace{-2.5em}
\end{figure}

\section{Simulations and Experimental Results}

Our experimental evaluation of the \ac{snnds} algorithm is organized into the following categories: 
\begin{enumerate}[label=\alph*)]
\item The performance of our algorithm is illustrated on a representative subset of the 2D LASA handwriting datasets\footnote{Publicly available at \url{https://bitbucket.org/khansari/lasahandwritingdataset}} as well as a 3D dataset\footnote{Available at \url{https://github.com/nbfigueroa/ds-opt}} in an environment with single obstacles, 
\item We provide a detailed performance analysis and benchmark comparisons of our algorithm with the \ac{abcds} algorithm~\cite{schonger_learning_2024} for the case of 2D handwriting datasets,
\item We evaluate the performance of demonstrations recorded kinesthetically by the Franka Emika Panda robots~\cite{panda_robot} in the presence of multiple obstacles. 
\end{enumerate}

\begin{figure}[t!]
\vspace*{1.0em}%
\centering
\includegraphics[width=1\columnwidth]{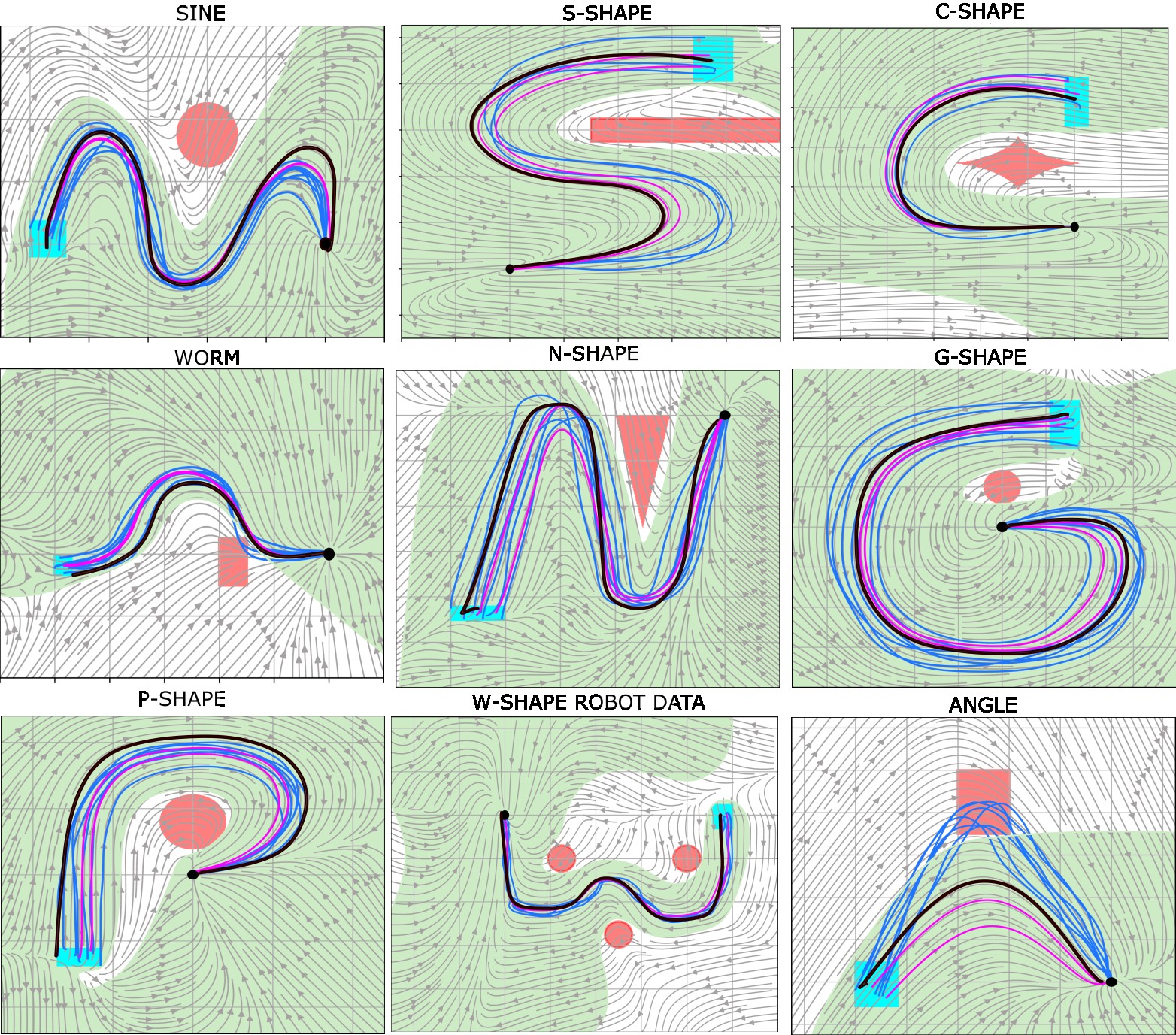}
\caption{Neural \ac{ds} generated by our proposed approach with obstacles for the LASA handwriting and robot demonstration datasets. Five demonstrations (blue) were used. The learned trajectories (pink) are simulated for two initial conditions within the initial set, and the robot path (brown) is obtained for another initial point. The region in green describes the safe set, while the arrows indicate the flow of the \ac{ds}.}
\label{fig:lasa}
\end{figure}

{\setlength{\parindent}{0cm}\paragraph{\textbf{2D and 3D Datasets}}
First, we consider $8$ datasets corresponding to different shapes in the LASA handwriting dataset, which is a standard benchmarking dataset used in the literature~\cite{figueroa-ds-gmm, schonger_learning_2024, nawaz_learning_2024}, and offers an simplification to several industrial tasks such as cutting, carving, and welding~\cite{welding}. The data are normalized to reside within $X= [-1,1]^2$, and without loss of generality, it is assumed that the attractor point coincides with the origin. The neural networks corresponding to $f_\theta, V_{\theta'}$ and $\B_{\theta''}$ are trained and certified using Algorithm~\ref{alg:nnds} with high confidence levels, specifically $1-\epsilon, 1-\beta \geq 0.99$. The simulation results presented in Fig.~\ref{fig:lasa} show multiple key advantages of the \ac{snnds} approach\footnote{The models used in the figures were obtained on an Ubuntu 20.04 LTS system with 16GB RAM equipped with NVIDIA GeForce RTX 4050 - 6GB GPU.}. From the evaluation: (\emph{i}) our approach can handle arbitrarily complex obstacle configurations without imposing any restrictions on the shape's convexity, star-shapedness, or semialgebraicity, as required in~\cite{dynamical_obst_avoid, schonger_learning_2024}. This is illustrated by a diamond-like obstacle that is considered for the C-Shape dataset. In particular, our algorithm only requires an analytical closed-form expression for the unsafe set. (\emph{ii}) \ac{snnds} can generate safe trajectories even with potentially unsafe demonstrations, as demonstrated by the Angle and Worm datasets. Here, the learned trajectories closely resemble the shape of the original demonstrations for all the initial conditions, though the dynamics have been modified to ensure safety. (\emph{iii}) the sub-zero-level set $\{x \in \R^n \mid \B(x) \leq 0\}$ of the barrier function, which defines the \emph{safe set}, outlined by the green regions in the figures, fit tightly between the obstacles and the demonstrations, showing that the certificates learned by our approach are non-conservative. As long as the robot lands in the green region when perturbed, safe navigation is formally guaranteed.}

\begin{figure}[t!]
\vspace*{1.0em}%
\centering
\includegraphics[width=0.6\columnwidth]{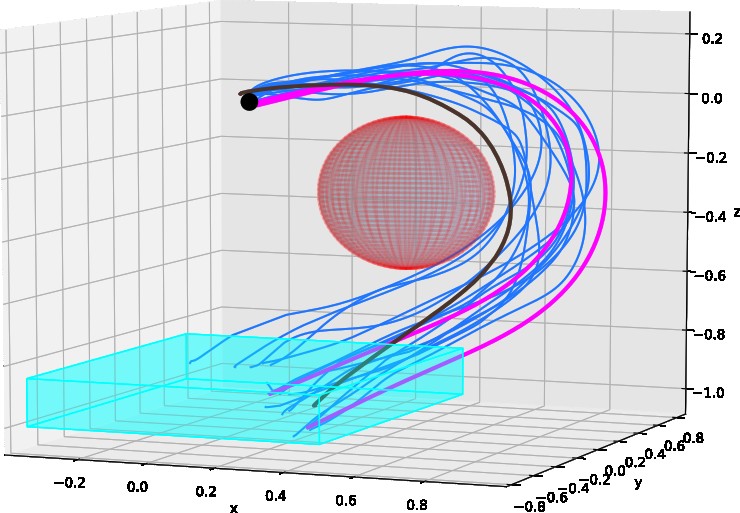}
\caption{Neural \ac{ds} generated by our proposed approach with obstacles for the 3D C-shaped motion. Ten demonstrations were used. The legend follows from Fig.~\ref{fig:lasa}.}
\label{fig:cshape}
\vspace{-\baselineskip}
\end{figure}

Furthermore, we extend our evaluation to a 3D C-shaped motion. Consistent with the 2D results, and with very high confidence levels of $1-\epsilon, 1-\beta \geq 0.99$, we find that the learned trajectories are not only safe and stable, but also closely mimic the demonstrations, as shown in Fig.~\ref{fig:cshape}. However, as is evident from the figure, a larger number of demonstrations are required to learn safe, generalizable trajectories due to the curse of dimensionality.

{\setlength{\parindent}{0cm}\paragraph{\textbf{Performance Analysis and Benchmark Comparisons}}
The performance of our S$^2$-NNDS algorithm is evaluated by measuring the mean, the standard deviation (SD) of the mean square error (MSE), and dynamic time warping (DTW) distance between the demonstrations and the learned trajectories (i.e., the value of our objective function~\eqref{eq:mse}),  on a set of test data corresponding to the LASA handwriting sets. While MSE measures the average squared, point-point distance between two trajectories, DTW measures the optimal alignment between two trajectories~\cite{dtw}. Therefore, MSE provides a good measure for analyzing how close the DS predicts motion w.r.t to the demonstrations, while DTW measures the similarity between the path shapes. First, we compare the results obtained in Fig.~\ref{fig:lasa} with ABC-DS in~\cite{schonger_learning_2024}, which characterizes the dynamical system as well as certificates as polynomial functions and solves a bilinear optimization problem [8] using PENBMI, a proprietary, commercial solver to solve bilinear problems. However, since we observed that PENBMI performs reliably only with specifically crafted semi-algebraic sets with at most one constraint (e.g. a hyper-rectangular initial set $a_i \leq x_i \leq b_i$, $i \in \{1,2\}$ requires four semi-algebraic constraints), we utilized ellipsoidal approximations for any non-ellipsoidal sets. The obtained results are provided in Table~\ref{tab:comparisons-ourexp}.\footnote{The ABC-DS results were obtained after executing the algorithm for a maximum of 100 iterations and validated with numerical tolerance (primal residual) of 0.001, as the code consistently terminated with a maximum iteration error.} Then, we also consider the LASA handwriting datasets and the obstacle configurations utilized in~\cite{schonger_learning_2024}. The obtained metrics are provided in Table~\ref{tab:comparisons-theirexp}.

The results show that the performance of our approach is competitive to ABC-DS. In cases involving highly nonlinear motion shapes with high rate of change in motion, and complex obstacle configurations, S$^2$-NNDS demonstrates superior performance with lower MSEs as well as DTW. Moreover, ABC-DS fails to provide any results when demonstrations are unsafe. 
Additionally, we observed that the S$^2$-NNDS algorithm may provide less conservative barrier functions, as illustrated in Fig.~\ref{fig:comp} and Table~\ref{tab:comparisons-area}, by the larger safe sets that lie tightly between the demonstrations and obstacles. However, a certain trade-off between MSE, DTW, and conservatism of the barrier function is reasonable. Note that, in general, ABC-DS can only handle semi-algebraic obstacle configurations, whereas our approach is more general and can handle non-semi-algebraic sets. 
For example, one cannot use diamond-like obstacles with ABC-DS, but our approach indeed handles this case well, as demonstrated by the $C$-Shaped motion in Fig. ~\ref{fig:lasa}. We also show a scenario in Fig. ~\ref{fig:sine_new} where \ac{snnds} succeeds and ABC-DS fails due to the fact that any semi-algebraic approximation (e.g. ellipsoid) leads to infeasibility as demonstrations become unsafe. Thus, \ac{snnds} is more suitable for complex obstacle configurations in tightly-spaced, cluttered environments.
However, note that while ABC-DS provides \textit{absolute} guarantees on \textit{global} asymptotic stability, S$^2$-NNDS only provides \textit{statistical} formal guarantees for \textit{local} asymptotic stability (c.f. Section~\ref{sec:disc}).   

\begin{table*}
\vspace*{1.0em}%
\centering
\resizebox{0.7\textwidth}{!}{
\begin{tabular}{ c|cccccccc }
 & Sine & S-Shape & C-Shape & Worm & Angle & G-Shape & P-Shape & N-Shape \\
 \hline

MSE: \ac{snnds}  & $0.015$ & $0.016$ &  $0.019$ & $0.026$  & $0.044$ & $ 0.019$  & $0.008$ & $0.084$  \\
MSE: ABC-DS  & $0.023$ & $0.014$ &  $0.029$ & $--$  & $--$ & $ 0.041$  & $0.027$ & $0.020$  \\
SD:  \ac{snnds}:  & $0.065$ & $ 0.063$ &  $0.077$ & $0.074$  & $ 0.109$ & $0.080$  & $0.045$ & $0.1937$  \\
SD:  ABC-DS:  & $0.0762$ & $ 0.061$ &  $0.096$ & $--$  & $--$ & $0.108$  & $0.0705$ & $0.081$  \\
DTW: \ac{snnds}: & $0.062$  & $0.234$ & $0.288$ & $0.071$ & $0.341$ & $0.737$ & $0.667$ & $0.157$ \\
DTW: ABC-DS: & $0.132$  & $0.336$ & $0.381$ & $--$ & $--$ & $0.8617$ & $0.95$ & $0.190$ 
\end{tabular}}
\caption{MSE, SD and DTW of the learned trajectories for the LASA Handwriting Datasets corresponding to Fig.~\ref{fig:lasa}, compared also with ABC-DS~\cite{schonger_learning_2024}. Note that ellipsoidal approximations for the initial and obstacle configurations were utilized for ABC-DS, and $--$ denotes cases where no satisfactory results were obtained.}
\label{tab:comparisons-ourexp}
\end{table*}

\begin{table*}
\centering
\parbox{.45\linewidth}{
\resizebox{0.45\textwidth}{!}{
\begin{tabular}{ c|cccccccc }
 & Sine & S-Shape & Worm  & P-Shape  \\
 \hline

MSE: \ac{snnds} & $0.035$ & $0.014$ &  $0.014$ & $0.016$   \\
MSE: ABC-DS & $0.025$ & $0.012$ & $0.066$ & $0.048$ &   \\
SD: \ac{snnds} & $0.11$ & $0.067$ &  $0.068$ & $0.06$  \\
SD: ABC-DS & $0.078$ & $0.060$ & $0.116$ & $0.109$  \\
DTW: \ac{snnds} & $0.0686$ & $0.119$ &  $0.068$ & $0.845$  \\
DTW: ABC-DS & $0.101$ & $0.521$ & $0.089$ & $0.532$  
\end{tabular}}

\caption{Comparisons of MSE, SD and DTW of the learned trajectories for the obstacle configurations in~\cite{schonger_learning_2024}.}
\label{tab:comparisons-theirexp}}
\vspace*{-0.2em}
\hfill
\parbox{.45\linewidth}{
\resizebox{0.45\textwidth}{!}{
\begin{tabular}{ c|cccccccc }
 & Sine & S-Shape & Worm  & P-Shape  \\
 \hline

Safe Area: \ac{snnds} & $2.02$ & $3.871$ &  $3.865$ & $3.59$   \\
Safe Area: ABC-DS & $2.68$ & $3.768$ & $3.101$ & $3.13$ \\  \end{tabular}}

\caption{Quantitative area comparisons of the safe regions generated by \ac{snnds} and ABC-DS for obstacle configurations used in~\cite{schonger_learning_2024}.}
\label{tab:comparisons-area}}
\vspace*{-\baselineskip}
\end{table*}
\vspace{-0.3em}

{\setlength{\parindent}{0cm}\paragraph{\textbf{Demonstrations from Franka Emika Robot}} To further evaluate the validity of our approach, we tested our approach on kinesthetically recorded demonstrations from the Franka Emika robot \cite{panda_robot}. To test the adaptability of \ac{snnds}, we introduced $3$ obstacles around the robot's intended path. Once again, the learned trajectories closely follow the demonstrations while avoiding the obstacles, as shown in Fig.~\ref{fig:lasa}.}  

{\setlength{\parindent}{0cm}\paragraph*{\textbf{Robot Experiments}}
To assess the practical feasibility of the proposed framework in a real setup, we utilize the robot drawing system depicted in Fig.~\ref{fig:ExperimentSetup} to draw shapes from the LASA dataset\footnote{The drawing robot in the drawing system is a 7-DoF Franka Emika robot \cite{panda_robot}, controlled via the Franka Control Interface at a frequency of 1 kHz. The control loop is executed on a computer equipped with an Intel Core i5-12600K CPU running Ubuntu 22.04 LTS with a real-time kernel version 5.15.55-rt48.}}
\begin{figure}[t]
\vspace*{1.0em}%
    \centering
    \includegraphics[scale = 0.05]{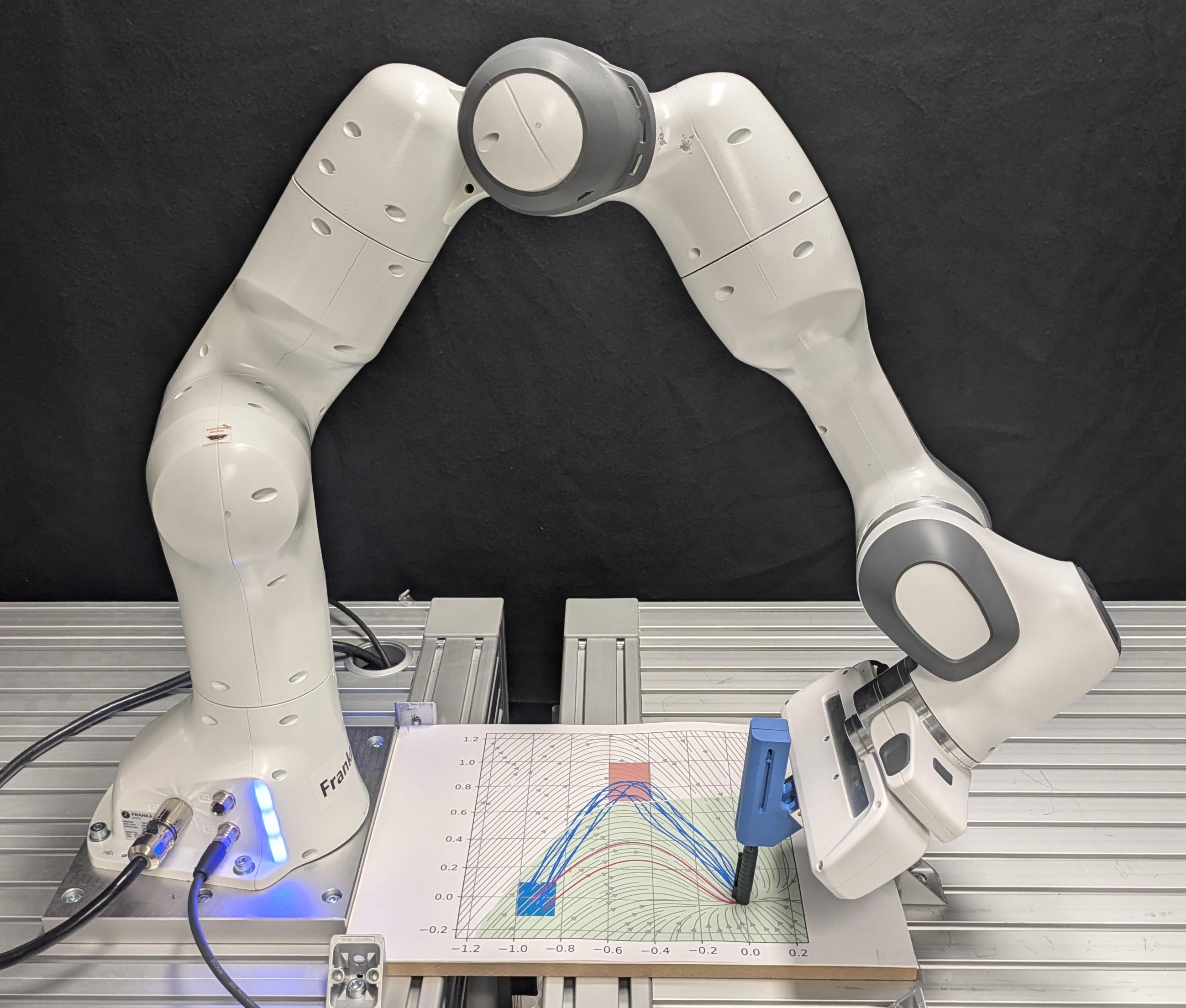}
    \caption{Drawing platform with Franka Panda Robot}
    \label{fig:ExperimentSetup}
\vspace{-\baselineskip}    
\end{figure}
The robot $Z$-axis and orientation are governed by an impedance controller to achieve consistent pen-strokes while preventing excessive force on the pen and paper. Additionally, passive interaction control from \cite{KronanderBillard2016PassiveInteractionControl} is utilized to follow the integral curves of the \ac{ds} generated by our method, enabling drawing along the $X$ and $Y$ axes while maintaining a passive relation between external forces and the robot velocity. The 3D C-shaped motion is also validated on the robot, with the translation axes being controlled by the DS generated by our method and orientation governed by an impedance controller. The resultant motion of the robot for the considered datasets (2D and 3D datasets) is recorded and plotted in Fig.~\ref{fig:lasa} and Fig.~\ref{fig:cshape}, respectively.

\begin{figure}
\vspace*{1.0em}%

\centering
\includegraphics[width = 0.8\columnwidth]{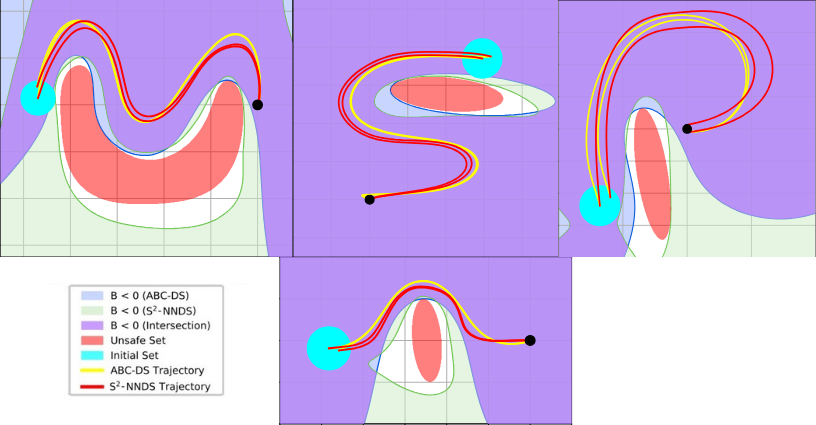}
\caption{Comparison of the learned trajectories and barrier functions obtained via \ac{snnds} and ABC-DS, respectively. Barrier functions learned via our approach generally fit more tightly around the obstacles and offer less conservative results, as can be seen from the area computed in Table~\ref{tab:comparisons-area}.}
\label{fig:comp}
\vspace{-0.1em}
\end{figure}

\begin{figure}
\centering
\includegraphics[scale = 0.4]{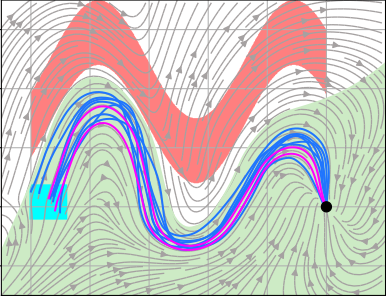}
\caption{Sine-like motion learned in a cluttered environment with a large sine-shaped obstacle. While \ac{snnds} produces reasonable motion, ABC-DS leads to infeasibility as any computationally tractable semi-algebraic approximation (e.g. ellipsoid) leads to an intersection between the demonstrations and the obstacles.}
\label{fig:sine_new}
\vspace*{-\baselineskip}
\end{figure}
\section{Discussion}
\label{sec:disc}
Despite the strong performance of S$^2$-NNDS for safe LfD, several limitations remain. Since NNs are trained on bounded domains, S$^2$-NNDS ensures only local asymptotic stability in a domain $X$, unlike prior work guaranteeing global stability~\cite{ds-gmm,schonger_learning_2024}. Thus, convergence to the origin is only guaranteed from some (unknown) region of attraction; starting or drifting outside this region may break convergence. Nevertheless, enforcing radial unboundedness in the Lyapunov function or careful NN tuning can yield strong empirical stability. Moreover, S$^2$-NNDS is trained offline and currently handles only static obstacles, although this still covers many practical scenarios such as warehouses. Finally, like most NN-based methods, its performance depends on hyperparameters and on the initial training of $f_{\theta}$. Addressing these challenges is left for future work.

\balance

\end{document}